# Field-Programmable Deep Neural Network (DNN) Learning & Inference accelerator: a concept

LUIZ M FRANCA-NETO*, Western Digital, Milpitas, CA, USA, luiz.franca-neto@wdc.com

ABSTRACT: An accelerator is a specialized integrated circuit designed to perform specific computations faster than if those computations were performed by general purpose processor, CPU or GPU. State-of-the-art Deep Neural Networks (DNNs) are becoming progressive larger and different applications require different number of layers, types of layer, number of nodes per layer and different interconnect between consecutive layers. A DNN learning and inference accelerator thus need to be reconfigurable against those many requirements for a DNN. It needs to reconfigure to maximum use its on die resources, and if necessary, need to be able to connect with other similar dies or packages for larger and higher performing DNNs. A Field-Programmable DNN learning & inference accelerator (FProg-DNN) using hybrid systolic/non-systolic techniques, distributed information/control and deep pipelined structure is proposed and its microarchitecture and operation presented here. 400mm$^2$ die sizes are planned for 100 thousand workers (FP64) that can extend to multiple-die packages. Reconfigurability allows for different number of workers to be assigned to different layers as a function of the relative difference in computational load among layers. The computational delay per layer is made roughly the same along pipelined accelerator structure. VGG-16 and recently proposed Inception Modules are used for showing the flexibility of the FProg-DNN's reconfigurability. Special structures were also added for a combination of convolution layer, map coincidence and feedback for state of the art learning with small set of examples, which is the focus of a companion paper by the author (Franca-Neto, 2018). The flexibility of the accelerator described can be appreciated by the fact that it is able to reconfigure from (1) allocating all a DNN computations to a single worker in one extreme of sub-optimal performance to (2) optimally allocating workers per layer according to computational load in each DNN layer to be realized. Due the pipelined nature of the DNN realized in the FProg-DNN, the speed-up provided by FProg-DNN varies from 2x to 50x to GPUs or TPUs at equivalent number of workers. This speed-up is consequence of hiding the delay in transporting activation outputs from one layer to the next in a DNN behind the computations in the receiving layer. This FProg-DNN concept has been simulated and validated at behavioral/functional level.

Appendices:
A.   CPLDs, FPGAs and FProg-DNN: parallel concepts.
B.   Convolution layer, Map coincidence and feedback: highlights.

Concepts: • **Field-Programmable** → Machine Learning acceleration; • **Datecenter** → large scale DNN learning & inference

## 1   INTRODUCTION

Field-Programmable Gate Arrays (FPGA's) use the high circuit densities in modern semiconductor fabrication processes to design integrated circuits that are "field-programmable". Their on-die logic is reconfigurable even after the dies are packaged and shipped. FPGA's use arrays of logic cells surrounded by programmable routing resources. Look-Up Tables (LUTs), Block RAM (BRAM), and specialized analog blocks (PLLs, ADC and DAC converters, and high speed transceivers with signal emphasis, continuous and decision feedback equalizers) are also added to high performance FPGA's. Interconnection resources dominate their logic resources and provide for the FPGA's flexibility (Farooq, 2012).

Field-Programmable DNN Learning & Inference Accelerator (FProg-DNN) presented in this work aims at providing for reconfigurable DNNs what FPGAs offer for reconfigurable logic. In FPGAs, unused hardware resources after synthesis, place and route, remain as idle hardware on die. In FProg-DNN, unused hardware resources will remain as idle tensor and pixel units.

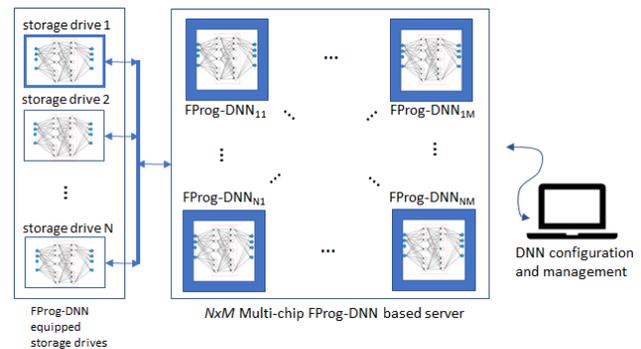

Fig. 1. Field-Programmable DNN (FProg-DNN) in datacenters: very large datasets are distributed among several drives and very large DNNs are realized in multi-chip FProg-DNN servers. Storage drives might also be equipped with smaller FProg-DNN chips for pre-screening subsets of datasets and diminishment of data traffic to server. User's configuration & management computer pushes model configuration files onto FProg-DNNs in server and onto each storage drive.

Figure 1 shows a system architecture envisioned for applications with large scale deployment of FProg-DNN chips. Many storage drives distributed in a datacenter or distributed across several datacenters are equipped with a relatively small version of a FProg-DNN in each storage unit. Very large datasets from several drives are brought to be processed by a much larger DNN realized in a server with multiple large size FProg-DNN chips.

The possibility of adding FProg-DNN to storage drives in addition to those FProg-DNN chips on the server is intended to

---

exploit the fact that the initial layers in a large DNN tend to produce much more activation outputs than later layers. These initial layers are also concerned with common low level edge patterns and might use the same trained coefficients among the storage drives. Moreover, this "pre-screening" by the local FP-DNN in each drive may bring the additional benefit of diminishing the amount of data (activation outputs) sent to the much larger multi-chip Fprog-DNN based server, thus leading to more efficient a computational solution.

FProg-DNN uses (a) reconfigurable functional blocks and (b) reconfigurable interconnect resources on die (appendix A). As per figure 1, a computer is used to define the architecture of a DNN to be synthesized. On this computer, the user specifies number of inputs, number of hidden layers what type of layers they are, number of nodes per layer and number of outputs. Each layer will have their non-linearity specified. All hyperparameters are also specified by the user for learning on the targeted FProg-DNN device. The computer than compile all this information and create a *model file* to be sent to each FProg-DNN in the server. Inside each FProg-DNN, a control processor (figure 2a) informs each worker its identity. Worker's identity defines the behavior of a worker, which kind of process it is to perform, what filters to use, how many map pixels to construct, and what data from the previous layer are relevant to its processing. Workers also have special functional blocks for analytics and diagnostics.

Plus, the user may define configuration parameter for the FProg-DNN inside each storage drive.

The physical interconnections that are used to bring data from those storage drives to the FProg-DNN server are used to provide model configuration from the user's computer to the FProg-DNN in each storage driver.

FPGA's are able to solve any problem that can be posed as a computing problem. That can be readily recognized by noting that a full-fledged soft processor can be implemented in an FPGA. Similarly, FProg-DNN can be reconfigured to any arrangement of convolution layers, max pool layer and fully-connected layer using its programmable functional blocks and interconnects. That is readily recognized as well by the fact that an FProg-DNN can be reconfigured (a) to make a single worker to run the computational load of all nodes in a DNN or (b) distribute worker among layers so that the delay from computations are roughly the same in each layer in the FProg-DNN.

The pipelined structure of the DNN realized in a FProg-DNN die is responsible for the speed-up in relation to GPUs or TPUs. Several DNN layers are synthesized in a FProg-DNN and the transport of activation outputs from one layer to the next is overlapped by the computation in the following layers. Thus, different from GPUs and TPUs, the delay in transporting data from one layer to the next is hidden in Fprog-DNN realizations.

## 2  FProg-DNN GENERAL ARCHITECTURE

In a simplified architectural description as shown in figure 2a, an FProg-DNN is defined by a large reconfigurable fabric and a control processor.

Inside the reconfigurable fabric, reconfigurable interconnect channels are alternated with reconfigurable functional blocks. As shown in figure 2b, these functional blocks are either tensor arrays fields or pixel arrays fields, which will be defined in more detail later.

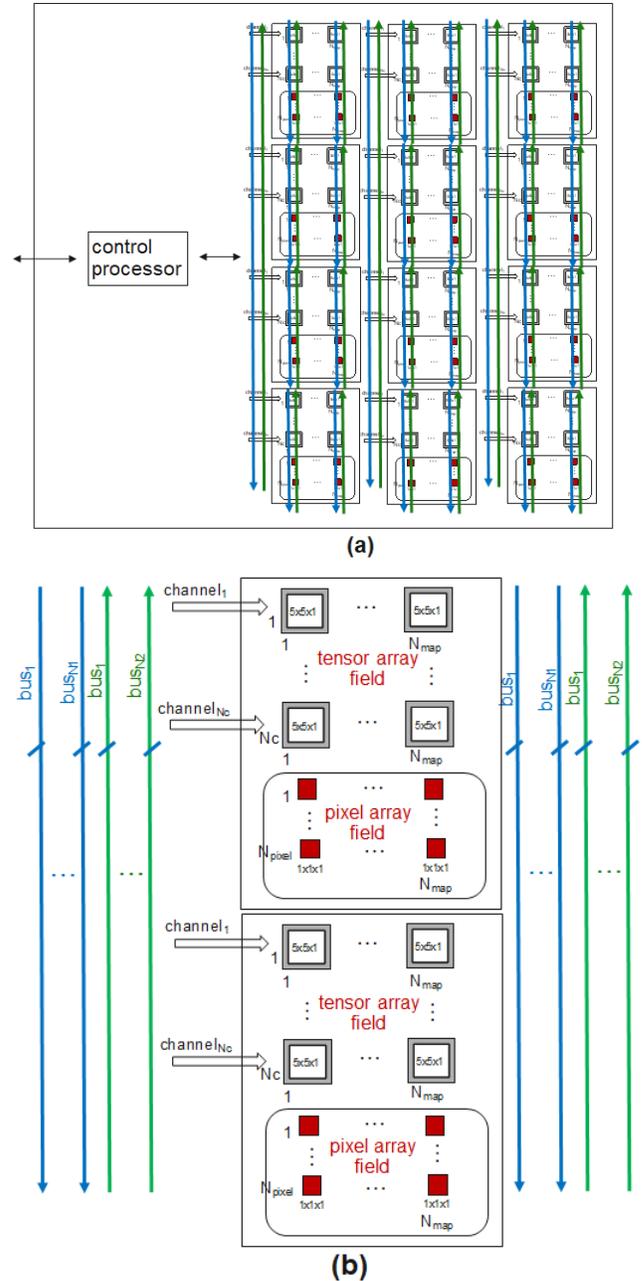

Fig. 2. Field-Programmable DNN Learning & Inference accelerator (FProg-DNN): (a) a control processor receives DNN model configuration and communicates to each worker its identity, i.e. what layer and nodes a worker represents, its computation scope, etc) and (b) the alternating reconfigurable interconnect channels, tensor array and pixel array fields on die. In a alternative design, the tensor array fields and pixel array fields are realized in different dies.

In the following sections, fundamental challenges and their solutions are discussed, and these solutions build up towards the final FProg-DNN microarchitecture.

### 2.1 VLSI wiring challenge in 1-node:1-worker mapping

A fully-connected layer with 1-node:1-worker mapping represents significant fan-out for the nodes producing activating outputs and also significant fan-in for the nodes receiving those signal in the following layer. Assuming $N$ nodes in each of the fully connected layers, this represent $O[N^2]$ number of wires. Using systolic transmission between those fully connected layers, this challenge becomes a much simpler $O[N]$ wire placement with shorter wires (less wires and shorter wires potentially leading to lower power operation as well).

Figure 3a and 3b shows the corresponding realization of a DNN with fully connected input and output layers plus two hidden layers. Only the forward propagation is illustrated.

Each network node in figure 3a is mapped to a *triplet* in figure 3b formed by node plus sourcing ($s^{[l]}$) and destination ($d^{[l]}$) systolic elements.

Activation outputs from the first hidden layer are transferred to the second hidden layer systolically using flow in both upward and downward directions. A cross-over connection splitting the number of nodes in each hidden layer by half speeds the transfer. Assuming $N$ nodes in each of the hidden layers, in $N$ systolic pulses, every one of the nodes in the second hidden layer will have seen the activation outputs from each one of the nodes in the first hidden layer.

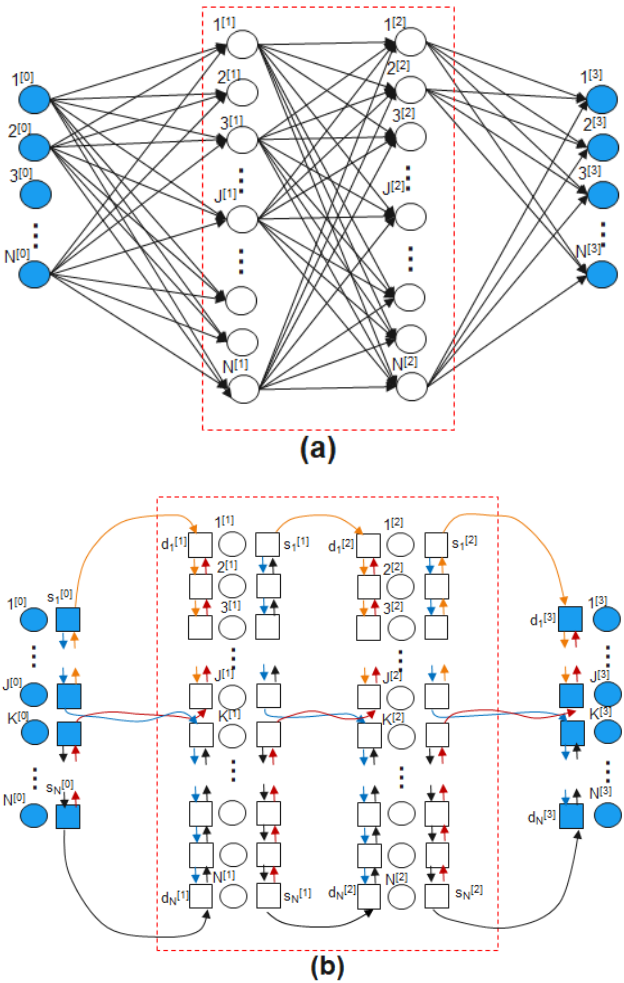

Fig. 3. Systolic transfer from one layer in the DNN to another. The large fan-out and fan-in $O[N^2]$ wiring placement in (a) is turned into point to point connections $O[N]$ wiring placement. $N$ being the number of nodes in fully-connected layer.

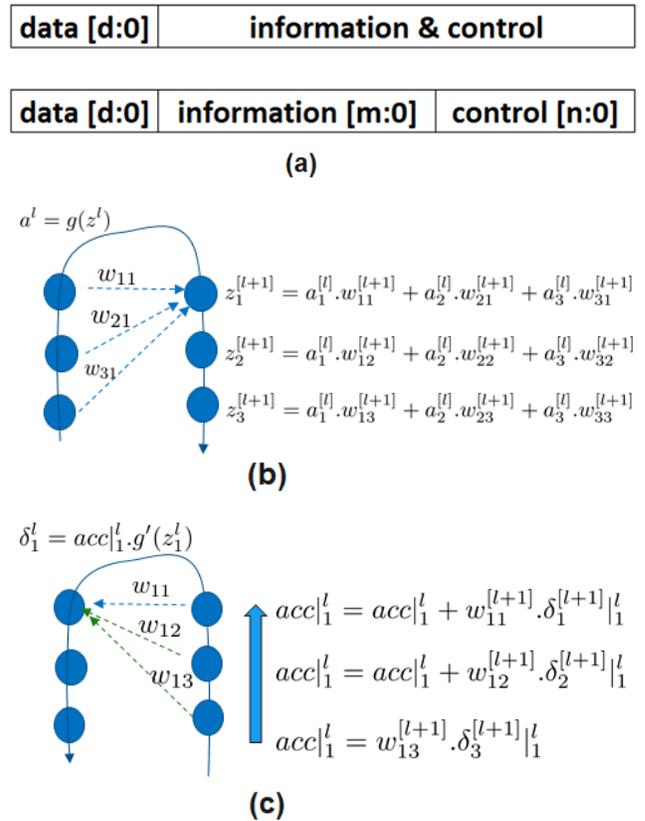

Fig. 4. Systolic transport of data plus information and processing control tag for forward and backpropagation. In forward propagation, every node in a layer reads the tag to identify activation outputs from the previous layer it needs to process. In backpropagation, the systolic flow of data and tagged information is reversed towards the previous layers. The partial derivatives of loss with respect to activation is backpropagated and accumulated along layer *[l+1]* as the accumulated results systolically moves back to the nodes in layer *[l]*.

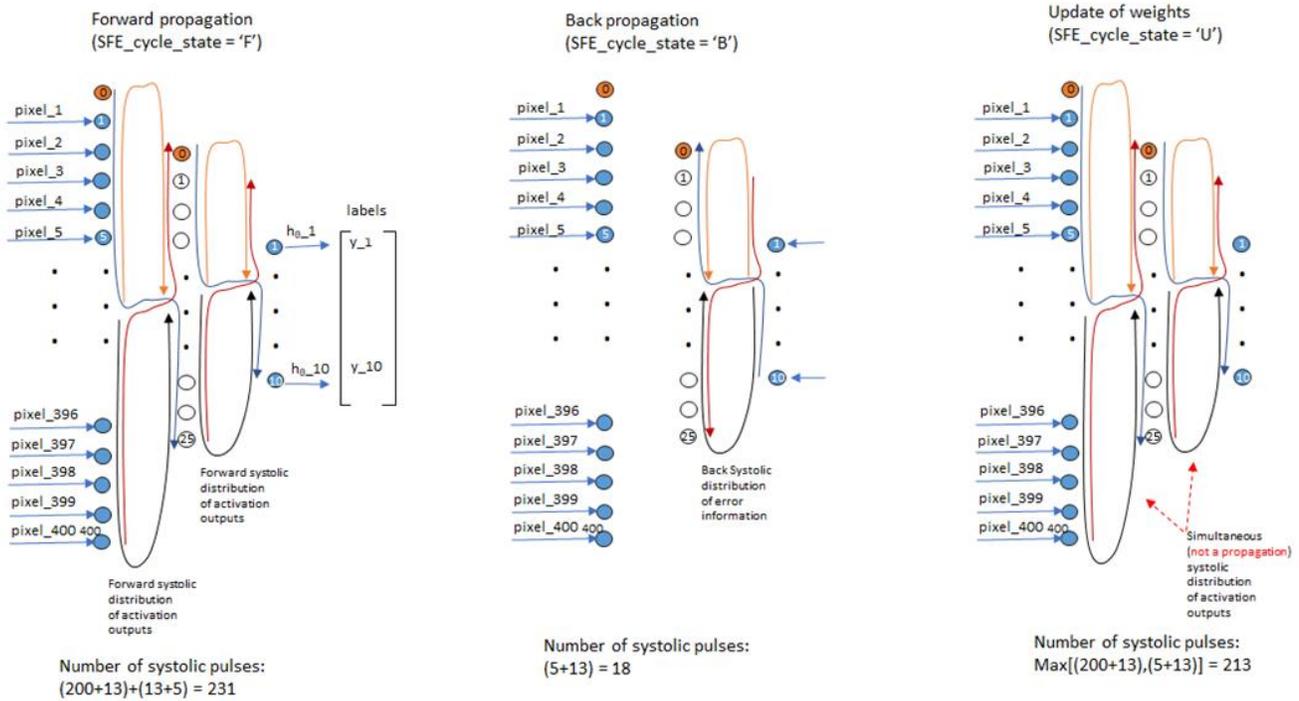

Fig. 5. Example of 2-layer fully-connected DNN with 400 input nodes, 25 hidden nodes and 10 output nodes for handwritten digit recognition. In training mode, the network will be sequentially in each of the three states: "F"orward, "B"ackpropagation, and "U"pdate [of weights]. The total number of systolic pulses involved in each state is readily calculated from the number of nodes involved as 231, 18 and 213 systolic pulses respectively.

Most importantly, each node in the second hidden layer can start the computation of its weighted sum of activation signals as soon as the first activation signal from the previous layer is received.

Each node can also finish its weighted sum of activation signals in $N$ systolic pulses.

Therefore, the delay in transferring the activation signals from one layer to the next is *hidden* as this delay is overlapped by the computation in each node in the receiving layer. This hiding of activation transport delay is responsible for the speed up against GPUs or TPUs realizations of the same DNN.

Additional cross-over paths can be added to speed up the transfer of activation outputs from one layer to the next.

Contrary to the activation signals that need to be routed to all nodes in the following layer, the weights used by the following layer are specific to each node in a fully-connected layer. Thus, systolic techniques are appropriately used in transferring activation signals, but weights might be directly transferred by non-systolic means to each node in fully-connected layers. Alternatively, weights can also be locally stored in each node. Later, it will be shown that in the interest of reconfigurability however, systolic transfer of data and weights are used extensively among tensor arrays and pixel arrays fields in a FProg-DNN die.

2.2 Tagging data for backpropagation and reconfigurability

Backpropagation and reconfigurability of interconnects in the FProg-DNN make use of information and control tags added to the data transferred between layers in the DNN. A data frame as used is shown in figure 4a.

In forward propagation, the identification of a node producing an activation signal is inserted in the tag for that activation signal. Because of the systolic transfer (fig. 3b) the activation signals reach each receiving node in different order. Each node however pairs (and record) weights and activation signal received for the computation of its weighted sum of activation signals as shown in figure 4b.

During backpropagation, when the systolic flow is inverted as in figure 4c, the recorded pairing of weighs and activation sources allows each node to properly pair its calculated partial derivative of activation outputs with respect to inputs ($\delta_i^{[l]}$) and the proper weight when backpropagating to a specific node in the previous layer. Different from the forward propagation, the backpropagation accumulates the weighted sum of partial derivatives as the data is propagated backwards. The tags inform the destination source in each node the passes through and each node adds its calculated partial derivatives and weights when destination addresses match.

| | VGG-16 | Activation Size | Act Function | Activation Count | Parameters to learn | Computational Load | | total # of workers (target): | | | |
|---|---|---|---|---|---|---|---|---|---|---|---|
| | | | | | | | | 100,000 | | | |
| | | | | | | | | # workers (equal computational load) | die area (%) | N (nodes/worker) | # of pixels/worker |
| | Input | (224,224,3) | | 150,528 | | | | | | | |
| 1 | Conv1 (f=3, s=1,same) | (224,224,64) | ReLU | 3,211,264 | 1,792 | 86,704,128 | | 560 | 0.56 | 5,730 | 89.53 |
| 2 | Conv2 (f=3, s=1,same) | (224,224,64) | ReLU | 3,211,264 | 36,928 | 1,849,688,064 | | 11,956 | 11.96 | 269 | 4.20 |
| 3 | MaxPool2 (f=2,s=2) | (112,112,64) | - | 802,816 | 0 | | | | | | |
| 4 | Conv3 (f=3, s=1,same) | (112,112,128) | ReLU | 1,605,632 | 73,856 | 924,844,032 | | 5,978 | 5.98 | 269 | 2.10 |
| 5 | Conv4 (f=3, s=1,same) | (112,112,128) | ReLU | 1,605,632 | 147,584 | 1,849,688,064 | | 11,956 | 11.96 | 134 | 1.05 |
| 6 | MaxPool4 (f=2,s=2) | (56,56,128) | - | 401,408 | 0 | | | | | | |
| 7 | Conv5 (f=3, s=1,same) | (56,56,256) | ReLU | 802,816 | 295,168 | 924,844,032 | | 5,978 | 5.98 | 134 | 0.52 |
| 8 | Conv6 (f=3, s=1,same) | (56,56,256) | ReLU | 802,816 | 590,080 | 1,849,688,064 | | 11,956 | 11.96 | 67 | 0.26 |
| 9 | Conv7 (f=3, s=1,same) | (56,56,256) | ReLU | 802,816 | 590,080 | 1,849,688,064 | | 11,956 | 11.96 | 67 | 0.26 |
| 10 | MaxPool7 (f=2,s=2) | (28,28,256) | - | 200,704 | 0 | | | | | | |
| 11 | Conv8 (f=3, s=1,same) | (28,28,512) | ReLU | 401,408 | 1,180,160 | 924,844,032 | | 5,978 | 5.98 | 67 | 0.13 |
| 12 | Conv9 (f=3, s=1,same) | (28,28,512) | ReLU | 401,408 | 2,359,808 | 1,849,688,064 | | 11,956 | 11.96 | 34 | 0.07 |
| 13 | Conv10 (f=3,s=1,same) | (28,28,512) | ReLU | 401,408 | 2,359,808 | 1,849,688,064 | | 11,956 | 11.96 | 34 | 0.07 |
| 14 | MaxPool10 (f=2,s=2) | (14,14,512) | - | 100,352 | 0 | | | | | | |
| 15 | Conv11 (f=3,s=1,same) | (14,14,512) | ReLU | 100,352 | 2,359,808 | 462,422,016 | | 2,989 | 2.99 | 34 | 0.07 |
| 16 | Conv12 (f=3,s=1,same) | (14,14,512) | ReLU | 100,352 | 2,359,808 | 462,422,016 | | 2,989 | 2.99 | 34 | 0.07 |
| 17 | Conv13 (f=3,s=1,same) | (14,14,512) | ReLU | 100,352 | 2,359,296 | 462,422,016 | | 2,989 | 2.99 | 34 | 0.07 |
| 18 | MaxPool13 (f=2,s=2) | (7,7,512) | - | 25,088 | 0 | | | | | | |
| 19 | FC14 | 4,096 | ReLU | 4,096 | 102,760,449 | 102,760,448 | | 664 | 0.66 | 6 | 6.17 |
| 20 | FC15 | 4,096 | ReLU | 4,096 | 16,777,217 | 16,777,216 | | 108 | 0.11 | 38 | 37.77 |
| 21 | Out | 1,000 | Softmax | 1,000 | 4,096,001 | 4,096,000 | | 26 | 0.03 | 38 | 37.77 |
| | | | total: | 15,237,608 | 138,347,843 | 15,470,264,320 | | 100,000 | 100.00 | | |

Fig. 6. VGG-16: mapping of N-node:1-worker is used to try to make equal delay in each layer as they are realized in FProg-DNN pipelined architecture.

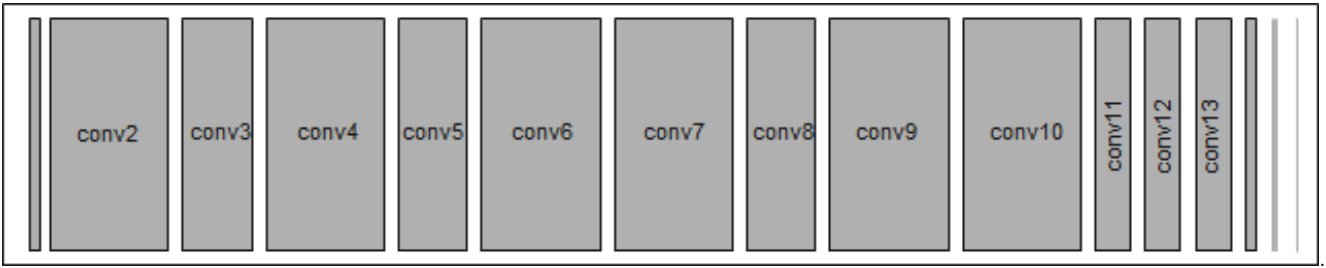

Fig. 7. VGG-16 in FProg-DNN: number of workers assigned to each layer for equal delay per layer in pipelined architecture indicates the relative area usage of each layer on-die.

The destination node in the previous layer will recognize its address in the tag and will take the companion data.

It is readily noticed from figure 4b-c that the backpropagation uses a transpose matrix of the weights used in forward propagation.

Figure 5 shows a DNN used for 10 handwritten digits recognition using 1 input layer with 400 nodes (20x20 size images), 1 hidden layer with 25 nodes and an output layer with 10 nodes (10 digit recognition). All delays can be represented in number of systolic pulses. The network can be trainded in Gradient Descent (GD), mini-batch GD or stochastic GD. The network is alternatively in "F"orward propagation, "B"ackpropagation or "U"pdating weights state. Additional control signals not shown in this illustration are used to place the network in each of those states. Those signals also control the state of propagation along the pipelined structure. Signals are used to indicate when activation signals are latched and ready for propagation to the next layer, other signals to indicate when all activation signals were transferred to the next layer.

## 3 FOLDING: moving towards N-Nodes:1-Worker mapping

State of the art DNNs may have millions of activation outputs (i.e. nodes) produced in one of its convolution layers. A *1-node:1-worker mapping* as used in the illustrations of Multilayer Perceptrons (MLPs) in the previous sections will not be usable. Hence, FProg-DNNs reconfigure their processing resources in *N-nodes:1-worker mapping*.

Figure 6 shows the VGG-16 deep neural network architecture. The number of activation outputs (nodes) in convolution layer-1 is more than 3 million.

Computational load is calculated for each layer using forward propagation as reference. Then a number of workers out of a pool of 100 thousand workers (on a 400mm$^2$ die for reference) is assigned to each layer so that the delay in completing the computational load in each layer is roughly the same. Figure 7 shows the effect of assigning workers as a function of computational load in each VGG-16 layer. The computational time in each layer is roughly the same and the size of the area of each layer on die reflects the number of worker allocated to each of those layers (a proxy, since workers in different layers can be quite different in size as discussed later).

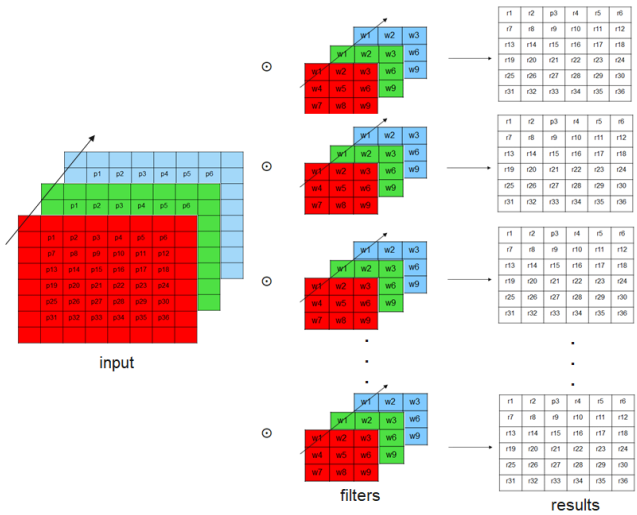

Fig. 8. Convolution layers: filters (kernels) use the same number of channels used in the input data. If many filters are used, many maps are produced as results.

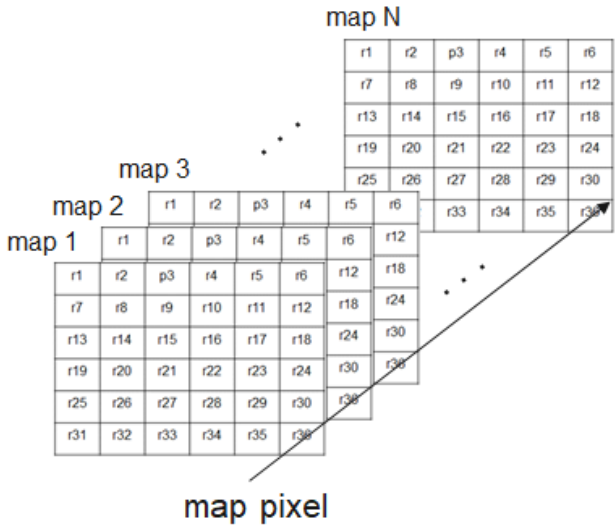

Fig. 9. "Pixel" definition: in this paper, an 1D array of results through the stack of maps produced by a convolution layer is defined as a "pixel". A worker in a convolution layer may be assigned to calculate the results for one or more pixels. If a worker is assigned to calculate ½, or another fraction of pixel, it will calculate that fraction of the depth of a pixel through the stack of maps.

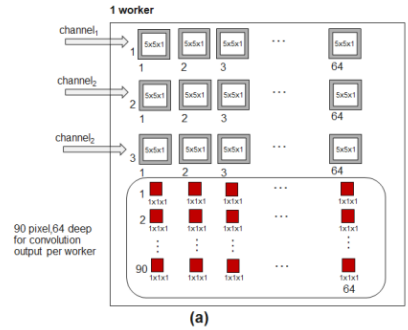

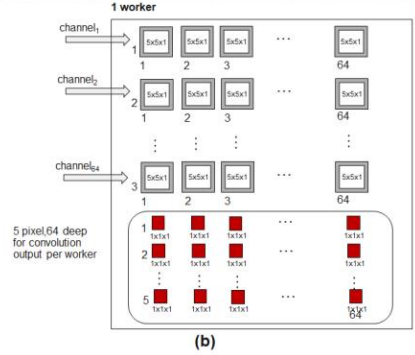

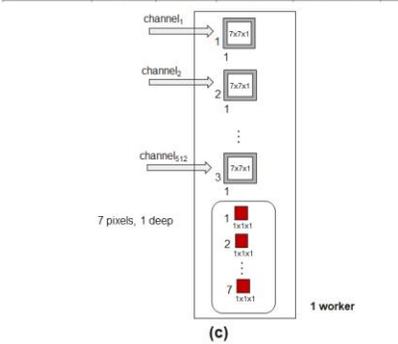

Fig. 10. VGG-16 in FProg-DNN: examples of layers from VGG-16 as synthesized by the tensor array and pixel array in FProg-DNN. (a) and (b) show different allocation of tensor array elements and pixel array elements. (c) shows a fully connected layer being realized as a special case of convolution layer. Data systolically progress from one 5x5x1 tensor multiply-accumulate to another in a row. Each tensor in a row sees different stride version of an image. Results from the tensor array in the same column (different maps) are added and after a non-linear function are stored in a pixel element. Data from the tensor array to the pixel array are tagged for proper routing from tensor array to pixel array, saving on wiring.

Convolution layers, as shown in figure 8 are used as a macro cell in FProg-DNN. MaxPool layers, Fully connected layers, as well as other much used layers are constructed as special cases of convolution layers.

Each worker is therefore assigned a number of "pixels" to calculate. A pixel is specifically defined in FProg-DNN as a 1D array of results across the stack of maps produced by a convolution layer as shown in figure 9. If a worker is responsible for a fraction of a pixel, this means it will produce only that fraction of results present in the corresponding pixel's 1D array.

### 3.1 VGG-16 layers in FProg-DNN

Figure 10a,b,c show how the final allocation of elements from the tensor array and pixel array to realize some of the layers in VGG-16. These final allocation, as it will be seen, might imply connecting more than one tensor array fields or pixel array fields to construct a worker.

### 4 COMPUTATIONAL SPEED-UP: HIDING THE DATA TRANSFER DELAY BETWEEN DNN LAYERS

As shown in figure 11, the pipelined structure of the DNN as realized in a FProg-DNN enables significant speed up compared to GPUs and TPUs.

That's because in GPUs and TPUs (Jouppi, 2017) the transfer of data from one layer to the next incurs in store and fetch of the activation outputs to/from a buffer/memory on die or external memory. In FProg-DNNs, the transfer of data from one layer to the next has a latency that is hidden behind the computation process of the receiving layer. Several layers of a DNN are pipelined on a FProg-DNN die and the deeper the pipeline, more significant the speed up. Assuming the same number of workers in either GPU/TPU and FPprog-DNN, the total time spent in computing while training or inferring will be the same, but the total time waiting for data will be much higher in GPUs and TPUs. This speed up can reach beyond 50x if store and fetches are to/from external DRAM. Moreover, due to its narrower width and pipelined structure, the workers in an FProg-DNN is more likely to be busy for a higher percentage of the time than GPUs or TPUs.

### 5 RECONFIGURATION OF ELEMENTS IN TENSOR ARRAY AND PIXEL ARRAY

General reconfigurability is enabled by the tensor array field and pixel array field in FProg-DNN and it is achieved by re-routing on-die resources. Rerouting in FP-DNN uses extensively tagging of data being transported. Tagging alleviates the wiring challenges as its akin to send data as *packetized* information. This same data tagging enables each worker to operate according to its identity and position in a layer as per the distributed information and control it reads from the data tag received.

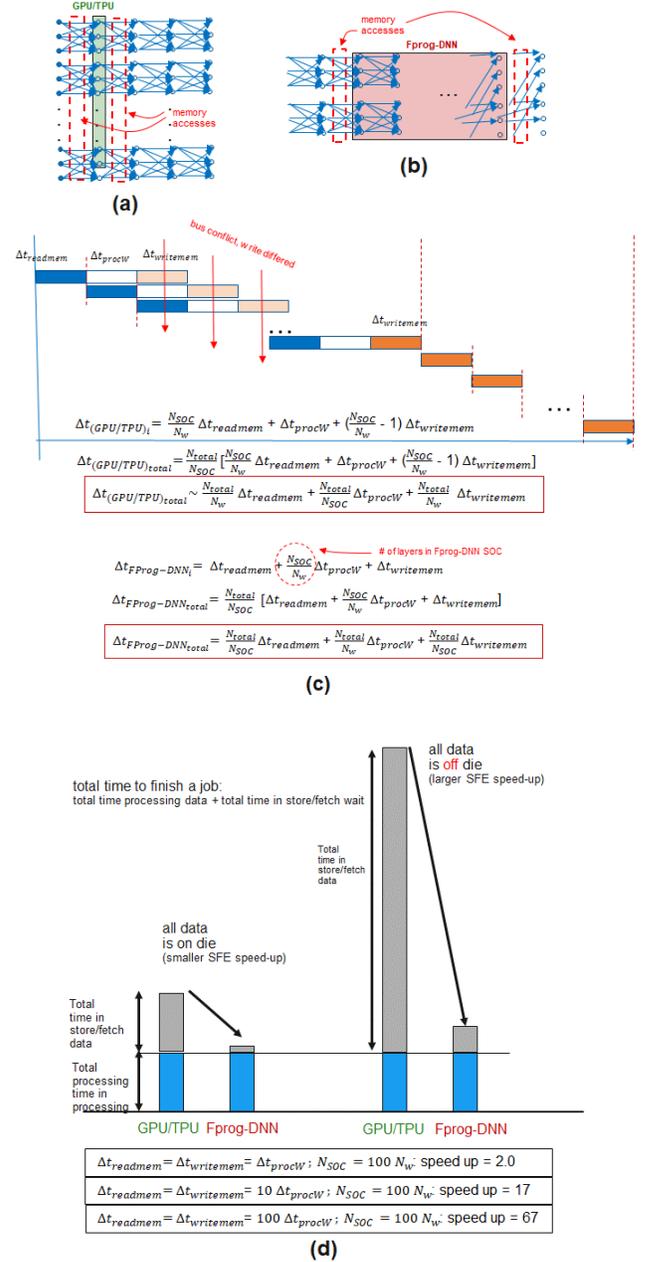

Fig. 11. Speed-up in FProg-DNN: (a) in GPUs and TPUs, every transfer of activation output from one layer to the next incurs in store and fetch data delays from on-die buffer or external DRAM. (b) FProg-DNN pipelined structure hides delay of transferring data from one layer to the next behind the computation delay in the receiving layer. (c) bus conflict while storing/fetching activation outputs worsens the case for GPUs and TPUs. FProg-DNN's speed-up can reach beyond 50x for store/fetches to/from external memory. $N_W$, $N_{SOC}$ and $N_{total}$ are number of nodes (or workers) per layer, in the die and in the larger DNN respectively. $\Delta t_{readmem}$, $\Delta t_{writemem}$, $\Delta t_{procW}$ are, correspondingly, time to read data for a layer from memory, time to write to memory data from one layer, and time to process data per layer.

A tensor element is defined as 5x5x1 multiply accumulate unit (figure 12a). Each pixel element has programmable non-linear function resources plus storage resources to support both forward and backpropagation local data. A worker in FProg-DNN is defined by allocation of tensor array elements and pixel array elements. In figure 12b, a worker needed more channels in its filters (kernels) than available in a single tensor array. Thus, additional elements from a subsequent tensor array field is added to define that worker. In that figure 12, red arrows indicate flow of data from the tensor array elements to the pixel array elements.

Similarly, if a worker is responsible for creating more pixels than available in a single pixel array field, additional pixels from a subsequent pixel array field are added to define that worker as shown in figure 12c.

If more maps are required for a worker, tensor and pixel elements more than one tensor array and pixel array fields are merged to define that worker.

The routing of information and the direction of information flow in both forward and backpropagation in tensor and pixel arrays follow the same principles of systolic movement of data and tagged information used in the inter-layer transport of activation outputs. Additional signals are used to indicate when to start a data transfer between array fields, and when a transfer in completed with data latched at the destination. This tagging is used to manage wiring complexity and support flexible reconfigurability.

### 5.1 Pruning and dropout

Pruning, the removal of irrelevant neurons after a training step as in (Le Cun et al. 1990), is supported by both zeroing the output of a irrelevant neuron and also signaling the pruning flag in the tag associated with that neuron output. This signaling in the tag will be used in the backpropagation. The DNN will progressively loose connections found irrelevant to the performance of the DNN and the modified network will be used during test.

The determination of which neuron is irrelevant can follow different algorithms. In the initial version of FProg-DNN, pruning is performed by the counting the number of times a neuron's output was zero or close to zero (by a hyperparameter defined by the user) during validation (therefore, after training and before testing). The final network after validation will therefore be different from the network defined initially by the user. Note that pruning can also be used at the convolution filter level too as in (Molchanov, 2017), and this is also supported by FProg-DNN.

Dropout is a random deactivation of neurons in a layer with probability *p*. Activation outputs for that layer are re-normalized by dividing those outputs by the keep probability, (1-*p*). In the FProg-DNN this adjust is merged with a specific enhancement algorithm for the activation outputs performed by an *Enhancement Matrix Unit* presented next.

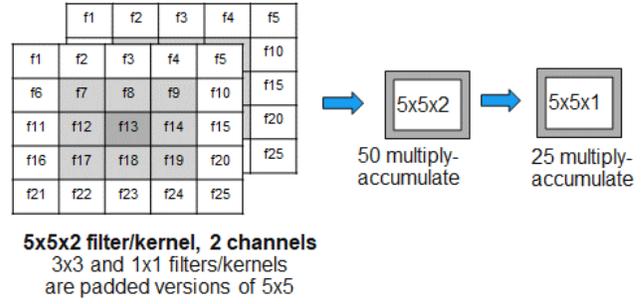

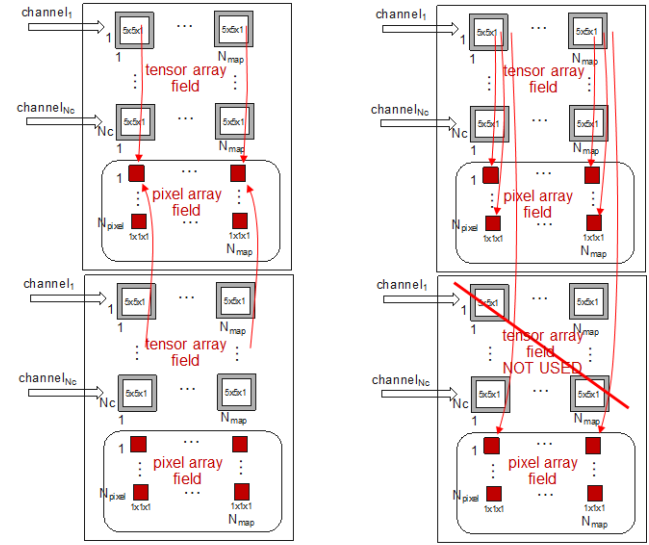

Fig. 12. Field reconfigurability: more than a single tensor array field or pixel array field might be combined to define a worker. Some resources from either field might stay unused. (a) a worker that uses filters with higher number of channels than provide in a single tensor array field is realized with elements from two tensor array fields. (b) a worker that produces more pixels than the number of pixels in a pixel array field is realized with elements from two pixel array fields. Red lines indicate the effects on routing computation results.

### 6 ALGORITHMIC INNOVATIONS: CONVOLUTION, MAP COINCIDENCE AND FEEDBACK

In (Sabour, 2017) the concept of capsules and *dynamic routing* between capsules is discussed as a means of providing richer representations of the data as patterns are found. This approach is also claimed to bring about better performance in identifying and recognizing highly overlapping objects.

A new VLSI-friendly approach to provide richer data representation referred to as "Convolution, Map Coincidence and Feeback" (ConvMapCoincidence+Feedback) is presented with results discussed by the author in a companion paper (Franca-Neto, 2018). This new approach is straightforwardly synthesized onto FProg-DNN.

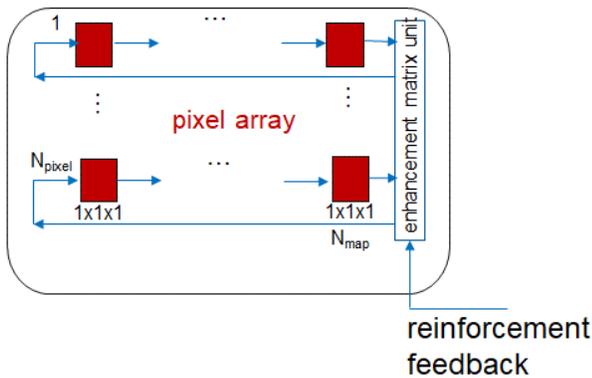

Fig. 13. Enhancement Matrix Unit: enhancement for map coincidence of activation outputs added to the pixel array field in FProg-DNNs.

Feedback is used in a pair of convolution layers to provide effect similar to dynamic routing in capsules (Sabour, 2017). In the approach discussed by the authors in the companion submission (Franca-Neto, 2018) this effect is achieved by the alignment of reinforcements of activation outputs between different convolution layers. In such a case, reinforcements at a lower level of pattern recognition is aligned with reinforcements in activation outputs at a higher level of pattern recognition. This also bring the additional benefit of maintaining state of the art performance with smaller datasets relative to DNNs without these activation output enhancements.

These algorithmic innovations are supported at the hardware level by the Enhancement Matrix Unit in the pixel array field of FProg-DNNs as shown in figure 13. Enhancements are applied to all the elements in 1D pixel array by a data-shifting procedure (figure 13).

### 6.1 Convolution, Map Coincidence and Feedback in Inception Modules

ConvMapCoincidence+Feedback added to Google's Inception modules (Szegedy, 2015) achieve better results in comparison to the standard use of those Inception Modules in accuracy and also in holding accuracy with smaller data examples.

As shown in figure 14, and Inception Module concatenate in a single layer outputs from 1x1, 3x3 and 5x5 convolution layers over the same input data from a previous layer. This arrangement of concatenated convolution layers provides alternate level of granularity in the detection of patterns in the same input.

Analysis of activity across the output maps created and concatenated by these layers provide rich reinforcement of relevant patterns with ConvMapCoincidence+Feedback layers. As shown in figure 14, ConvMapCoincidence in a Inception Module is constructed by adding the Enhancement Matrix Unit after the concatenation filter of those Inception modules. Two Inception Modules in a DNN are thus modified and connected by the reinforcement feedback connection.

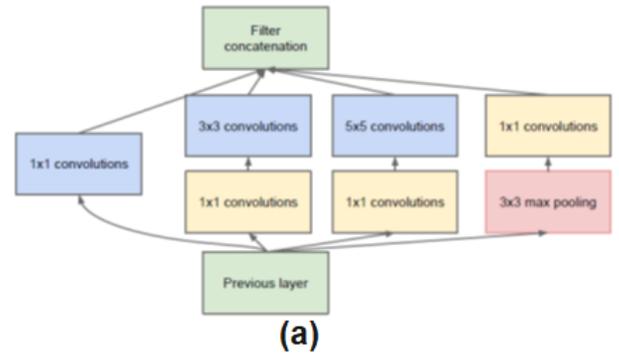

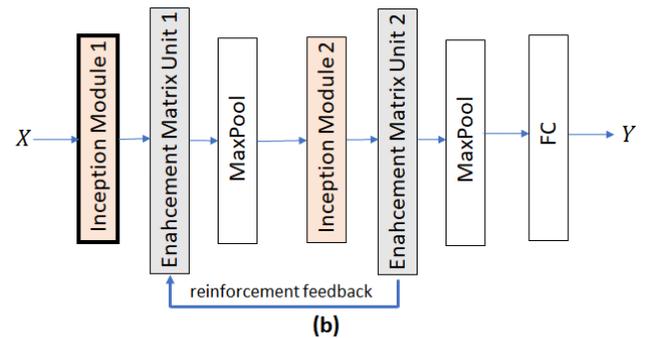

Fig. 14. ConvMapCoincidence+Feedback layers in Inception Modules: enhancement layers are added after the concatenation layers of inception modules and a feedback connection between a pair of inception modules is used to further reinforce activation output from both inception modules.

Appendix B provide additional details on this new algorithmic approach using convolution, map coincidence and feedback.

## 7 CONCLUSIONS

A Field-Programmable DNN Learning & Inference Accelerator, FProg-DNN, was presented and discussed in this paper. It was shown to be a flexible and capable solution for reconfigurable DNNs acceleration. State of the art very large DNNs tend to repeat the same fundamental blocks in an attempt at managing computational complexity. This trend provided further opportunities to use specific building blocks like 5x5x1 tensor multiply-accumulate elements as functional building blocks. Convolution layer structures are then used as generic layer for all the other layers in a DNN. A special new algorithm and ConvMapCoincidence+Feedabck layer enhancement tailored to state-of-the-art performance in learning from reduced number of examples was introduced with more details discussed in a companion paper by the author (Franca-Neto, 2018). This algorithm is straightforwardly realized with the support on elements in this FProg-DNN. Because of the pipelined structure in FProg-DNN, delays in transferring activation outputs from one layer to the next are hidden behind the computation delay in the following layer. The speed-up in FProg-DNNs relative to GPUs and TPUs can reach beyond 50x due to the benefit of this hidden transport delay. This FProg-DNN concept has been simulated and validated at behavioral/functional level.

A. CPLDs, FPGAs, and FProg-DNN accelerators: parallel concepts

Complex Programmable Logic Devices (CPLDs), Field-Programmable Gate Arrays (FPGAs) and Field-Programmable Deep Neural Network (DNN) learning & inference (Fprog-DNN) accelerators share two parallel conceptual constructions:

(a) suitable reconfigurable functional blocks, and
(b) suitable reconfigurable interconnection network on die

Figure A.1 shows the *suitable reconfigurable functional blocks* of CPLDs, FPGAs and FProg-DNN accelerators. Xilinx's original products are used here as representative examples of CPLDs and FPGAs. CPLDs inherit the NAND structure with inputs and inverted inputs used in Programmable Logic Arrays (PLAs). FPGAs use look-up tables (LUTs). LUTs allowed FPGAs to realize much larger number of different functions in comparison to CPLDs. A $k$ input LUT can realize $2^{2^k}$ different functions. FProg-DNN accelerator use tensor array fields and pixel array fields tiled throughout a die. The non-linearity specified for each layer is programmed in the pixel array field, which store activation outputs from a given worker and also partial results necessary for backpropagation. It's also possible to have two-die stack, where the tensor array fields are tiled in one die and the pixel array fields are tiled in a second die.

Programmability is realized with the use of SRAM cells in the Xilinx's illustrative examples. Those SRAM cells are connected in a long chain in the chip so that they can be programmed by scanning in a long stream of bits. FPGAs are thus programmed by a *bitstream* file.

Figure A.2 shows *the suitable reconfigurable interconnection networks on die* for FPGAs and FProg-DNN accelerators. FPGAs surround their reconfigurable functional blocks, named Configurable Logic Blocks (CLBs) by Xilinx, with Connection Boxes (CBs) and Switch Boxes (SB) to realize their reconfigurable interconnection network. Reconfigurability is programmed into SRAM cells as it is also done for the reconfigurable functional blocks, CLBs. Segments of different lengths are available on die for performance optimization of placing and routing (fig. A.2.a).

According to figure A.2.b, FProg-DNN accelerator use (1) pass transistors programmed by SRAM cells much like FPGAs and (2) semi-packetized interconnection network. In (2), interlayer transport of activation outputs (in forward propagation) or partial derivatives (in backpropagation) uses data tagged with information and control bits. Each worker is equipped with intelligence to recognize its position and role in a DNN layer. Hence, each worker recognize from the interlayer traffic, which data is for it to process. Interlayer traffic between tensor array fields and pixel array fields is also based on similar semi-packetized concepts. The pixel array fields are responsible for programmable non-linearities and store both forward and backpropagation related data.

*Field-Programmable DNN Learning & Inference accelerator*

Due to the expected progressive augment of complexity with generations of FProg_DNN accelerators, it's anticipated that automatic electronic design tools need to be developed. In figure A.3, an earlier vision for the necessary jobs to be implemented in Electronic Design Automation (EDA) tool for FProg-DNN is suggested vis-à-vis EDA tasks in use for FPGAs.

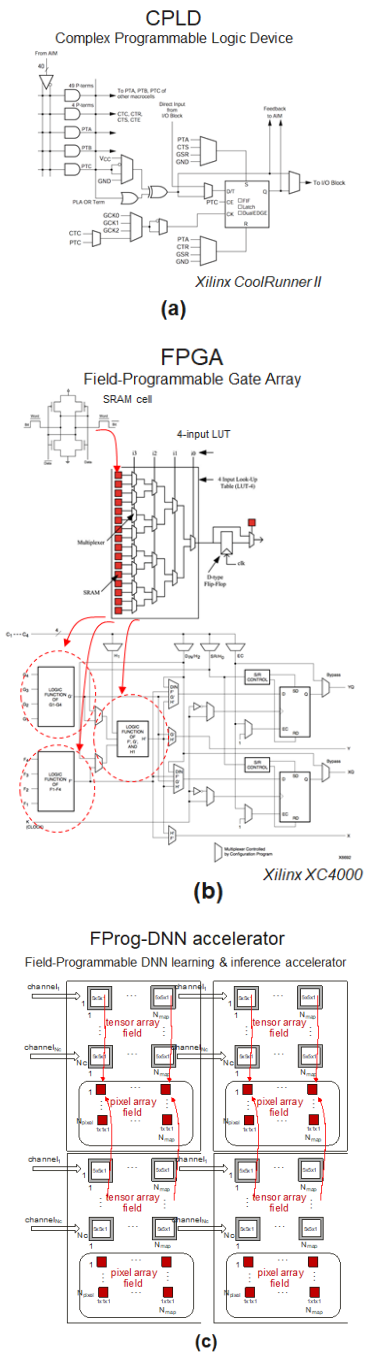

Fig. A.1. suitable reconfigurable functional blocks: (a) Complex Programmable Logic Devices (CPLDs) re-visit the NAND with inputs and inverted inputs structure used previously in Programmable Logic Arrays (PLAs). (b) Field-Programmable Gate Arrays (FPGAs) preeminently use Look-Up Tables (LUTs) (Ref.: U. Farooq et al, *Tree-Based Heterogeneous FPGA Architectures*, Springer, NY, 2012). (c) Field-Programmable DNN learning and inference accelerators use tensor array fields and pixel array fields which are tiled over a die.

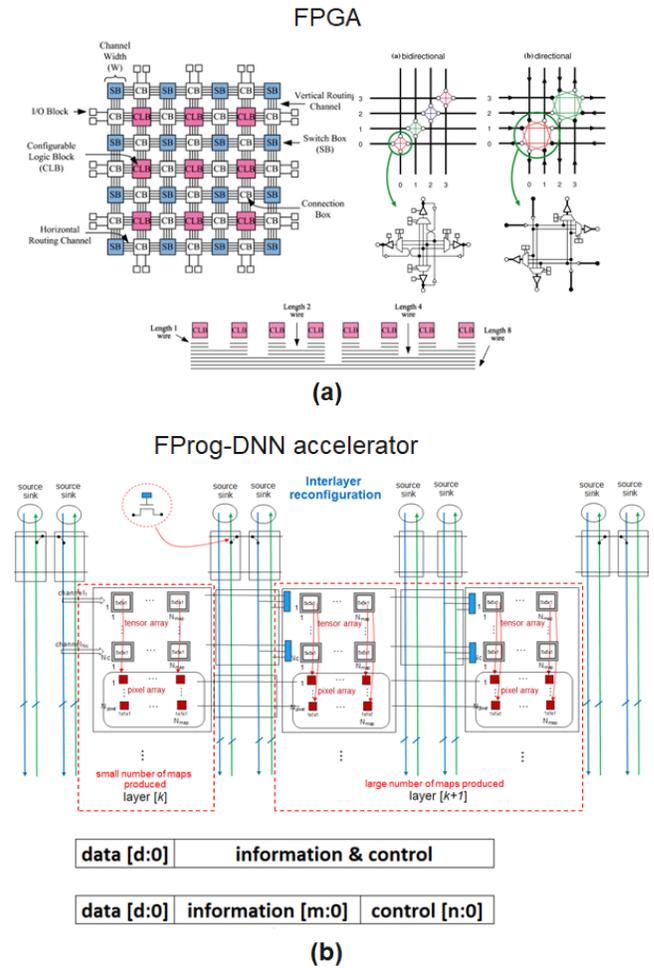

Fig. A.2. suitable reconfigurable interconnection network on die: (a) FPGAs surround their Configurable Logic Blocks (CLBs) with Connection Boxes (CBs) and Switch Boxes (SB) (Ref.: U. Farooq et al, *Tree-Based Heterogeneous FPGA Architectures*, Springer, NY, 2012). (b) FProg-DNN accelerators use pass transistors for wire connection and information plus control tags to transmitted data for a semi-packetized programmable interconnection network switching.

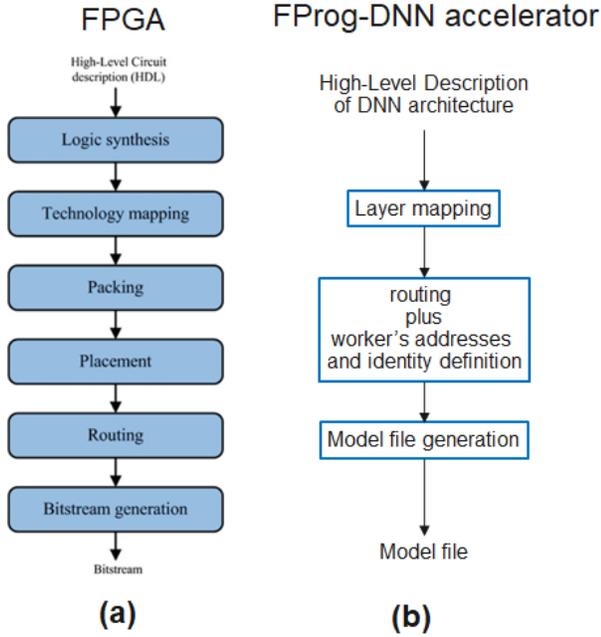

Fig. A.3. Electronic Design Automation (EDA): (a) FPGA EDA flow (Ref.: U. Farooq et al, *Tree-Based Heterogeneous FPGA Architectures*, Springer, NY, 2012) and (b) FProg-DNN accelerator.

B. Convolution layer, Map coincidence and feedback: highlights

As more detailed analysis is presented in a companion paper by the author (Franca-Neto, 2018), only highlights are visited in this appendix. Enhancement of a convolution layer activation outputs is implemented by the following *Enhancement Matrix Unit* used in FProg-DNN accelerator as per figure B.1.

This enhancement tends to exclude irrelevant features in the image. For instance, if the images have a crack in the background wall, that crack might be detected by one or a few of the filters used by the convolution layer. But, likely, no other filter might become active around that location of the crack in the image.

A following convolution layer will detect higher level patterns in the image. The output from this deeper DNN layer will also be enhanced by its corresponding Enhancement Matrix Unit following that convolution layer.

Feedback from the Enhancement Matrix Unit at the deeper convolution layer is used to further enhance the coefficients in the Enhancement Matrix Unit in the previous convolution layer. This reinforcement feedback used the additional connection in the FProg-DNN accelerator. The new activation outputs are sent to the subsequent convolution layer and the final outputs from this deeper convolution layer are now set to propagate forward in the DNN.

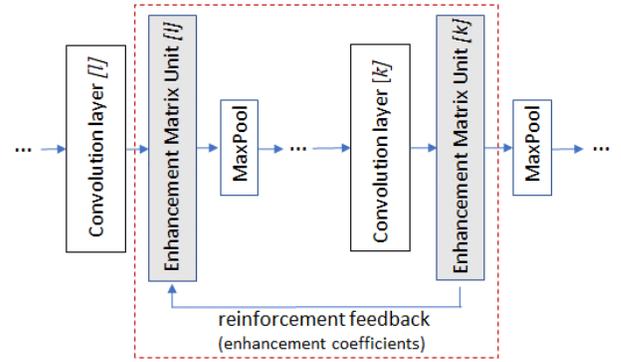

Fig. B.1. Enhancement and single-iteration feedback: in a first pass, enhancement coefficients calculated by the Enhancement Matrix Unit in layer *[l]* are applied to the convolution layer *[l]* prior to MaxPool. When data reaches convolution layer *[k]*, enhancement coefficients are calculated for this layer and also fedback to the Enhancement Matrix Unit in layer *[l]*. This Matrix Unit *[l]* multiplies element-wise the original enhancement coefficients with these fedback coefficients (assuming convolution outputs with same width and height). Data is then propagated to convolution layer *[k]*, new enhancement coefficients are calculated and applied at that layer and output from convolution layer *[k]* is finally sent forward to deeper layers in the neural network. This is referred to as a "single loop iteration" operation for the map coincidence enhancement.

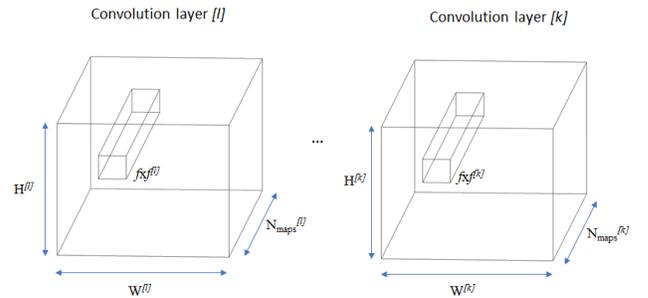

Fig. B.2. Mask for map coincidence: mask *fxf* can be 1x1, 3x3, 5x5 and 7x7. The sum of the values or magnitudes of the activation outputs encompassed by the mask volume produces one coefficient in the *enhancement matrix*. The strides are sized to avoid overlapping between elements of neighbor mask volumes. Width (W) and height (H) in convolutions *[l]* and *[k]* are the same. Number of maps $N_{map}[l]$ and $N_{map}[k]$ may differ.

This single loop iteration with the two convolution layers' Enhancement Matrix Units is used to establish baseline performance for the technique. DNN applications however might benefit significantly by a higher number of iterations, or by the use of feedback from more than one deeper convolution layer, or the use of feedback between convolution layers further apart in the DNN.

The further reinforcement by the feedback tends to enhance low level patterns that also appear around similar areas in the maps where higher level patterns are also detected. These reinforcements by coincidence across different levels of pattern recognition in a DNN are in effect similar to the "dynamic routing" in (Sabour, 2017) for capsules.

It is nevertheless noticed that the reinforcement used in this present work operates at the level of nodes and not at the level of capsules (collection of nodes in a layer per Sarbour, 2017). Moreover, the approach used in this work and choices among possible strategies were influenced by the author's interest in maintaining a straightforward path to realization in the hardware described for the FProg-DNN accelerator.

As indicated in Figure B.2, each enhancement coefficient in the Enhancement Matrix Unit is calculated by summing over the activation outputs (or the magnitude of those activation outputs) in a mask-defined volume across the maps created by a convolution layer. Masks of size 1x1, 3x3, 5x5 and 7x7 might be used to define the volume of clusters of activation outputs considered in the summation. Strides are correspondingly 1, 3, 5 and 7 pixels to avoid overlapping elements from neighbor mask-defined volumes. The summation of all the activation outputs (or magnitudes only) implied by each mask through all the maps represent the level of geographical coincidence in pattern detected by the different filters used in a convolutional layer. Each mask volume produces one coefficient to be placed in a matrix named *enhancement matrix*. If the mask used is 1x1, 3x3, 5x5, or 7x7, the enhancement coefficients calculated are applied to all the content in each activation output inside a 3D volume implied by one of those masks. The matrix of enhancement coefficients is produced by the *Enhancement Matrix Unit* as shown in figure B.1.

All the summation results in the enhancement matrix are "softmaxed" so that the no enhancement coefficient is larger than "1" and all the sum of the coefficients add up to "1".

The objective in this feedback of enhancement coefficients is to reinforce *coincidences in geographical location* of activation outputs across *different levels of pattern recognitions*. If, for instance, the shallower convolution layer detects and enhances activation outputs related to "noses" and "eyes", and the deeper convolution layer detects and enhances activation outputs related to a "face", the feedback will make more likely that only those activation outputs related "noses" and "eyes" at positions consistent with the "face" detected will be further reinforced. Or, in other words, an isolated "nose" or "eye" detected in the shallow convolution layer that can't be part of a face (perhaps, they are part of an abstract painting) will tend to be de-enhanced or suppressed in a DNN intent on detecting and recognizing actual faces.

This enhancement method is readily extended to Inception Modules as shown in figure B.3 and B.4 by recognizing that the inception modules concatenate convolution outputs. As in the previous discussion, the reinforcement feedback might connect two consecutive convolution layers or convolution layers further apart.

Mask size, number of iterations in the feedback loop, number of feedback loops and separation between convolution layers connected by a reinforcement feedback link imply different levels of computation complexity, implementation complexity in a FProg-DNN and latency for output results.

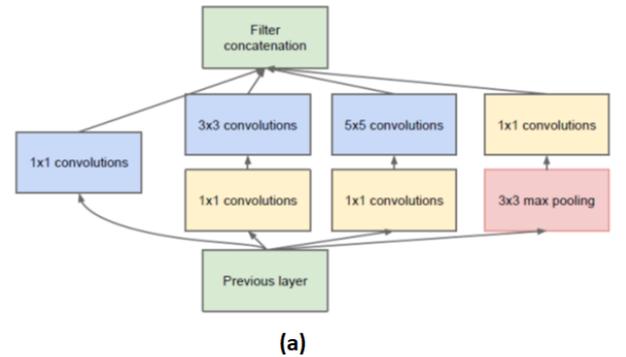

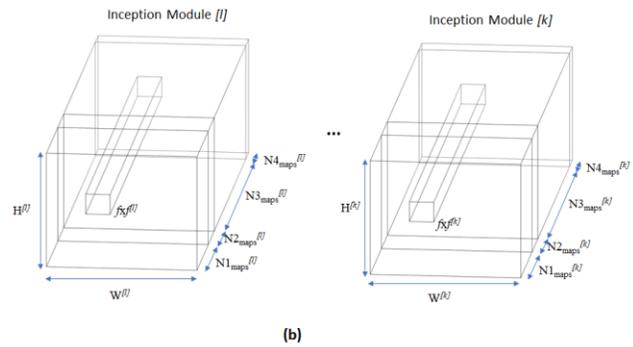

Fig. B.3. Mask for map coincidence in Inception Modules: the concatenation in the same layer of maps of 1x1, 3x3, and 5x5 convolution outputs from the same input data augments the probability map coincidence plus feedback brings more significant benefits.

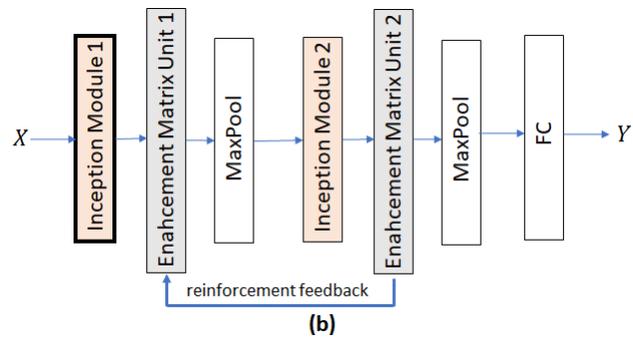

Fig. B.4. Adding Enhancement Matrix Units to Inception Modules